\documentclass[runningheads]{llncs}

\usepackage[T1]{fontenc}

\usepackage{graphicx}
\usepackage{amsmath}  
\usepackage{amssymb}
\usepackage{enumitem}
\usepackage{natbib}
\usepackage[ruled,vlined,linesnumbered]{algorithm2e}
\newcommand{\Pinc}{$\gamma$-monotonic }
\newcommand{\tnorm}{\mathcal{T}}

\usepackage{listings}
\usepackage{tikz}
\usetikzlibrary{shapes.geometric, arrows.meta, positioning, fit, backgrounds}
\usepackage{subcaption}

\newcommand{\nome}{\mbox{\rm FLASP}}

\newcommand{\TT}{T_\Pi}
\newcounter{proprule}
\renewcommand{\theproprule}{P\arabic{proprule}}

\begin{document}

\title{Applying Answer Set Programming with Fuzzy Membership Functions: a Case Study}
\titlerunning{\nome: Answer Set Programming with Fuzzy Membership Functions}

\author{Luca Ferragina\inst{1}\orcidID{0000-0003-3184-4639} \and
Ilenia Galati\inst{1}\orcidID{0009-0006-3566-7296} \and
Lorena Gullone\inst{1}\orcidID{0009-0004-8142-0513}
\and
Francesco Scarcello\inst{1}\orcidID{0000-0001-7765-1563}}


\institute{DIMES Dept., University of Calabria, 87036 Rende (CS), Italy.\\
\email{\{luca.ferragina,ilenia.galati,lorena.gullone,francesco.scarcello\}@unical.it}}

\maketitle  

\begin{abstract}
Human reasoning often operates through qualitative concepts expressed by linguistic labels such as \textit{high}, \textit{low}, \textit{expensive}, or \textit{cheap}, whose interpretation depends on context and is usually vague, despite being rooted in numerical data. This paper explores a novel fuzzy-logic-based qualitative extension of Answer Set Programming (ASP) to bridge numerical information and qualitative reasoning. The underlying language, formally introduced in a separate work, provides a principled framework that avoids rigid thresholds and supports robust reasoning under vagueness.

Focusing on a representative use case, we illustrate how the framework integrates numerically grounded inputs (such as outputs of machine learning models) with symbolic reasoning over qualitative labels. Key features, including learning-based membership functions and semantically enriched predicates, enable the combination of expert knowledge, contextual factors, and subjective interpretations within a unified declarative setting.

\keywords{Fuzzy Logic \and
  Answer Set Programming \and
  Neuro-symbolic AI \and
  Membership Functions}
\end{abstract}

\section{Introduction}

Recent advances in machine learning have significantly enhanced the ability to extract patterns from unstructured data. However, despite their predictive accuracy, such systems often remain difficult to interpret and audit, particularly in applications where decisions must be traceable and grounded in explicit assumptions. In contrast, symbolic reasoning frameworks offer transparency and formal rigor, but they typically struggle to capture the continuous, noisy, and context-dependent nature of real-world information. This gap becomes especially evident when expert knowledge is expressed through qualitative notions (e.g., \emph{high}, \emph{low}, \emph{severe}, \emph{acceptable}) that are inherently vague, even when derived from numerical measurements. Traditional approaches often rely on rigid thresholds to map numerical values to qualitative categories, leading to brittle behavior in which small variations in input may cause abrupt and unintuitive changes in outcomes.

Fuzzy logic offers a natural foundation to address this issue by enabling reasoning with graded truth values. In this work, we build on a novel qualitative, label-based fuzzy extension of Answer Set Programming (ASP), introduced in a companion paper, which explicitly couples linguistic labels with numerically grounded information through membership functions. In this framework, predicates are enriched with sets of qualitative labels, and their truth degrees are constrained both by data-driven membership functions (for input predicates) and by semantic compatibility conditions (for derived predicates), expressed through admissible truth profiles.
Let us illustrate this point through a rule extracted from a program in a travel recommendation scenario:

\[
\begin{split}
\mathtt{recommend}(X, C; \textit{affordable}) \leftarrow & \mathtt{client}(C)\ \odot\mathtt{price}(F, C; cheap) \ \odot \\ & \mathtt{flight}(X, F) \ \odot \mathtt{hotel}(X, H) \odot \\ & \mathtt{price}(H, C; cheap) 
\end{split}
\]

The predicate \texttt{recommend} has labels \textit{infeasible} and \textit{affordable}, while \texttt{price} is described by \textit{cheap}, \textit{medium}, and \textit{expensive}. The rule assigns to \textit{affordable} a degree bounded by the aggregation (via a t-norm $\odot$, e.g., $\odot_{\min}$) of the body, thus reflecting that affordability depends on both flight and hotel being sufficiently cheap.

Note this notion is vague, and the mapping from actual money values to the degree of \textit{cheapness} is provided by suitable membership functions associated with the predicate \texttt{price}.
Such mapping can be determined by those studies that relate the actual perceived value of money for different individuals, typically depending on their annual income.
That is, the notion of \textit{cheap} is not universal, but depends on contextual and subjective factors: $\$200$ is a medium price for a rich customer, but it is very expensive for a low-income customer. Thus, in this fuzzy program, economic status is treated not as a passive filter but as a semantic dimension that adapts the concept definition.
In particular, the same travel package can therefore receive different graded valuations of \textit{affordability} from different users.
 
This example reflects a general setting where numerical information (often produced by statistical or learning methods) is integrated into a logical framework, allowing experts to express rules naturally while retaining control over graded outcomes.

It is worthwhile noting that our language is a direct extension of classical, two-valued ASP, and it naturally permits the definition and manipulation of crisp predicates and rules. Whenever there are only crisp predicates, all  fuzzy operators collapse to their classical counterparts, namely conjunction, disjunction, and negation, and we recover precisely Answer Sets semantics.

\medskip
The focus of this paper is on illustrating the practical effectiveness of this approach in a concrete setting, namely damage detection in the production line of screws, arising from a collaboration with an industrial partner. In such environments, automated visual inspection systems produce numerical features (such as defect size, shape irregularity, or color intensity) derived from image processing and machine learning components. These features must then be interpreted in qualitative terms that align with domain expertise, such as assessing whether a defect is \emph{small}, \emph{moderate}, or \emph{severe}, and whether a screw is \emph{repairable} or should be \emph{discarded}.

Our framework enables this transition by associating each qualitative label with a membership function that maps numerical observations to degrees of membership, thus preserving the inherent vagueness of the domain. On top of these input predicates, higher-level concepts are defined through declarative rules, while admissible truth profiles ensure that the resulting qualitative assessments remain semantically coherent. This allows the system to detect inconsistencies when conflicting evidence arises, and to propagate partial information across related qualitative dimensions.

We illustrate how this approach supports a full reasoning pipeline for quality control: from low-level perception, through qualitative abstraction, to decision-making actions such as \emph{discard}, \emph{rework}, or \emph{accept}. The example highlights how numerically grounded data and symbolic rules can be seamlessly integrated, yielding outcomes that are both flexible and aligned with domain-specific interpretations. While the formal properties of the language are proved in the companion work, this paper emphasizes its role in enabling a robust and expressive modeling of real-world decision processes in a concrete use-case.

The main contributions of this paper are: (i) the instantiation of the proposed qualitative fuzzy ASP framework in a realistic industrial damage-detection scenario, and (ii) the design and prototyping of a complete reasoning pipeline integrating data-driven membership functions with rule-based qualitative inference.


\section{Related Works}

The integration of uncertainty in rule-based reasoning has evolved from quantitative extensions of logic programming to neuro-symbolic approaches. Early works \cite{van1986} introduced attenuation factors, while Fuzzy Logic Programming (FLP) \cite{vojtavs2001fuzzy,ebrahim2001,medina2001,achs1995fuzzy} adopted a truth-functional view. These ideas were later extended to structured formalisms such as Description Logics and Answer Set Programming (ASP).

Handling uncertainty with non-monotonic negation has been studied in generalized logic programs. Parametric Deductive Databases were extended in \cite{loyer03,loyer09} with approximate well-founded semantics and lattice-based frameworks. In parallel, multi-adjoint logic programming was enriched with non-monotonic reasoning \cite{cornejo18,cornejo20}, providing expressive semantics and translations to simpler fragments.

In fuzzy Description Logics \cite{straccia2001reasoning,bobillo2008fuzzydl,lukasiewicz2008managing}, vagueness is modeled via membership functions while preserving decidability. In ASP, fuzzy extensions \cite{alviano2013fuzzy,janssen2009general} generalize stable models using $[0,1]$ truth degrees. CFASP \cite{Janssen12} showed that many constructs can be reduced to a core language. Further studies addressed negation and inconsistency \cite{blondeel2013fuzzy,van2006fuzzy}, introducing notions such as unfounded-freeness. Fuzzy Linguistic Logic Programming \cite{le2009} instead performs reasoning directly over linguistic terms. More recently, fuzzy Datalog with existential rules \cite{lanzinger2024fuzzy} achieved high expressivity while retaining tractable data complexity.
Neuro-symbolic AI complements these developments by combining learning and reasoning. \textit{DeepProbLog} \cite{deepproblog} and \textit{NeurASP} \cite{neurasp}, along with \textit{SLASH} \cite{slash}, integrate neural outputs into probabilistic or ASP-based reasoning, while Logic Tensor Networks \cite{badreddine2022logic} embed logical constraints into differentiable frameworks.
Note that all these approaches clarify, from different perspectives, how non‑monotonic reasoning, uncertainty management, and graded truth can coexist. However, they do not consider membership functions and qualitative labels as main components of the semantics of programs, which includes dealing with the implicit constraints based on how membership functions of qualitative labels may co‑vary.

\section{Preliminaries}

In this section, we introduce the core elements of our language, including its syntax, the fundamental components of its semantics, and the main algorithm underlying the inference process.

\subsection{Syntax of \nome}
A program $\Pi$ in \nome\ includes a set $\mathcal{C}$ of constants, a set of $\mathcal{V}$ variables, and a set $\mathcal{P}$ of predicate symbols.
Constants and variables play the same role as in standard logic programming.
Predicates, however, are enriched with additional structure to capture linguistic terms and fuzzy information.

As usual in Datalog, we distinguish two classes of predicate symbols in $\mathcal{P}$:
\begin{itemize}
  \item \emph{Input} (or \emph{extensional}, EDB) predicates, whose truth
        degrees are given as inputs to the program;
  \item \emph{Derived} (or \emph{intensional}, IDB) predicates, whose truth
        degrees are obtained by rule-based inference within the program.
\end{itemize}
We write $p(\mathbf{x};\ell)$ for the atom obtained from $p$, an argument tuple
$\mathbf{x}$, and a label $\ell\in\Lambda_p$, and we read its truth degree as the
\emph{degree of membership} of $\mathbf{x}$ in the fuzzy concept named by $\ell$.
Each predicate symbol $p^{(k)} \in \mathcal{P}$ of arity $k$ is associated with a triple $\langle \Lambda_p, \mathrm{M}_p, \Sigma_p \rangle$, where $\Lambda_p$ is a finite set of \emph{linguistic labels} describing the linguistic domain of $p$, $\mathrm{M}_p$ is a family of \emph{membership functions}, and $\Sigma_p$ is a collection of \emph{S-norms}.

For each label $\ell \in \Lambda_p$, the set $\Sigma_p$ provides an associated S-norm $\mathcal{S}_\ell : [0,1] \times [0,1] \rightarrow [0,1]$, which satisfies commutativity, associativity, monotonicity, and the boundary condition $\mathcal{S}_\ell(x,0)=x$. Typical examples include the Gödel S-norm, defined as $\mathcal{S}_{\max}(x,y)=\max(x,y)$, and the probabilistic sum, defined as $\mathcal{S}_{\mathrm{prob}}(x,y)=x+y-x\cdot y$.
The S-norm $\mathcal{S}_\ell$ aggregates the contributions to the truth/membership value for label $\ell$ from the rules defining atoms over predicate $p$, while membership functions are used to determine the degree to which a given value satisfies a linguistic label.

\paragraph{Membership Functions.} For an EDB predicate $p$, the family $\mathrm{M}_p$ contains, for each label $\ell \in \Lambda_p$, a membership function $\mu_\ell : \mathbf{D} \rightarrow [0,1]$, where $\mathbf{D}$ denotes the cartesian product of the domains of the underlying variables.
Each function $\mu_\ell$ assigns to a tuple $\mathbf{x} \in \mathbf{D}$ a degree of membership in the fuzzy set associated with the label $\ell$.
Membership functions are fixed \emph{a priori}, for instance on the basis of domain expertise or by means of machine learning techniques applied to data \cite{medasani}.
The resulting membership degree $\mu_\ell(\mathbf{x})$ is interpreted in the standard fuzzy-set-theoretic sense: $\mu_\ell(\mathbf{x})=0$ denotes non-membership, $\mu_\ell(\mathbf{x})=1$ denotes full membership, and intermediate values express partial membership (see Figure~\ref{fig:mf_kmeans}).

It is convenient to view $\mathrm{M}_p$ as a single function
$\mathrm{M}_p:\mathbf{D}\to[0,1]^{\Lambda_p}$, where each component supplies the result
for one label. The set of all compatible label-vectors,
\[
  \gamma_p \;:=\; \mathrm{M}_p(\mathbf{D}) \;\subseteq\; [0,1]^{\Lambda_p},
\]
is precisely the image of $\mathrm{M}_p$, and plays for EDB predicates the same
role as the admissible truth profile defined below for IDB predicates: it
records which combinations of label-truth degrees can actually occur.

For a derived predicate IDB $p$, 
we would like to compute membership values by means of program rules, according to the intended semantics of the program.
However, not every combination of truth values for the labels are logically meaningful, thus for IDB predicates we are interested in the membership-values compatibility set $\gamma_p$.
With this respect, it is convenient to represent it in a parametric way, with respect to a parameter
$t$, so that $\gamma_p$ is a mapping
\[
  \gamma_p:[a_p,b_p]\to[0,1]^{\Lambda_p},
  \qquad
  \gamma_p(t)=\bigl(\gamma_p(t)_\ell\bigr)_{\ell\in\Lambda_p},
\]
where the \emph{closed interval} $[a_p,b_p]$ is the parametrization domain, and $\forall t \in [0,1]$, $\gamma_p(t) = (\gamma_p(t)_\ell)_{\ell\in\Lambda_p}$ represents a coherent assignment of truth degrees to all labels of $p$.
Intuitively, $\gamma_p$ describes the intended semantic structure of the predicate $p$: each label $\ell \in \Lambda_p$ corresponds to one dimension of the truth space, and the curve $\gamma_p$ encodes how these labels may consistently co-vary.

\paragraph{Terms, literals, rules, programs.} A \emph{term} is either a constant $c \in \mathcal{C}$ or a variable $x \in \mathcal{V}$.
A \emph{fuzzy literal} is an expression of the form $p(\mathbf{x};\ell)$ or $\neg_{\scriptscriptstyle \mathcal{N}} p(\mathbf{x};\ell)$, where
$p \in \mathcal{P}$ is a predicate symbol of arity $k$,
$\mathbf{x}=(x_1,\dots,x_k)$ is a tuple of terms,
and $\ell \in \Lambda_p$ is a linguistic label associated with $p$.\footnote{In the examples, we omit the label if there is only one label and it is irrelevant, for instance for crisp EDB predicates.}
The former is referred to as a (positive or \emph{directed}) atom, while the latter as a \emph{negated} atom.
Formally, a negated atom is denoted as $\mathcal{N}(p)$ where $\mathcal{N}:[0,1] \rightarrow [0,1]$ is a decreasing function that satisfies $\mathcal{N} (0) = 1$ and $\mathcal{N} (1) = 0$. Typical negator examples include the \textit{\L{}ukasiewicz Negator} $\mathcal{N}_L (x) = 1 - x$ and the \textit{Gödel Negator} $\mathcal{N}_G (x) = 1 \text{ if } x=0$ or $\mathcal{N}_G (x) = 0 $ otherwise.

A \emph{fuzzy Datalog rule} is an expression of the form
\[
h \leftarrow b_1 \odot_{\tnorm} \cdots \odot_{\tnorm} b_m,
\]
where the head $h$ is a (positive) fuzzy atom and the body consists of a (possibly empty) conjunction of fuzzy literals $b_1,\dots,b_m$.
The operator $\odot_{\tnorm}$ denotes conjunction interpreted via some T-norm $\tnorm$.
Intuitively, the body specifies a set of conditions whose satisfaction contributes to the truth value of the head.

For a rule $r$, denote by $H(r)$ its head atom and by $B(r)$ the set of literals in its body where $B^+(r)$ and $B^-(r)$ denote, respectively, the sets of positive and of negated atoms occurring in the body of $r$.
Then, consider a (ground) atom $h$, and denote by $R_h$ the set of rules having $h$ as their head.

We assume the existence of a reserved predicate $\mathit{const}(x;\ell_c)$, where $x$ is some constant term. It is used to encode constant truth values within rules and
is equipped with the membership function $\ell_c(x)=1$, assigning truth value $1$
to any given term $x$. We write $\mathit{const}(v;\ell_c)$ to inject the numeric truth
value $v$.

The logical conjunction \textsc{and} is modeled by a T-norm $\tnorm : [0,1] \times [0,1] \rightarrow [0,1]$, which satisfies commutativity, associativity, monotonicity, and the boundary condition $\tnorm(x,1)=x$.
Commonly used examples include the Gödel T-norm $\tnorm_{\min}(x,y)=\min(x,y)$,
the product T-norm $\tnorm_{\mathrm{prod}}(x,y)=x\cdot y$,
and the \L{}ukasiewicz T-norm $\tnorm_L(x,y)=\max(0,x+y-1)$.

A \emph{fuzzy (Datalog) program} $\Pi$ is a finite set of rules.
A \emph{fact} is a rule with empty body. 
The \emph{grounding} of a program $\Pi$ is obtained by replacing all variables occurring in $\Pi$ with constants from the domain.
A program is said to be \emph{ground} if it contains no variables.
The \emph{Herbrand domain} $H_C$ of a program $\Pi$ is the set of all constants appearing in $\Pi$.
The \emph{Herbrand base} $\mathcal{B}_\Pi$ is the set of all ground fuzzy atoms that can be constructed from the predicate symbols in $\mathcal{P}$ and the constants in $H_C$.

\subsection{Semantics}

A \emph{fuzzy interpretation} $\mathcal{I}$ is a function 
$
\mathcal{I} : \mathcal{B}_\Pi \rightarrow [0,1]$
that assigns a truth degree to each ground atom $p(\mathbf{C};\ell)$ in the Herbrand base $\mathcal{B}_\Pi$.
The value $\mathcal{I}\bigl(p(\mathbf{C};\ell)\bigr)$ is interpreted as a truth degree of the atom with respect to the label $\ell$, or as a degree of membership to the fuzzy set $\ell$.
As we discuss below, classical (crisp) atoms can be modeled as atoms with a membership function having binary values only.

Denote by $p(\mathbf{C};\cdot)$ the basic unlabeled version of any ground atom with over predicate $p$ and with terms $\mathbf{C}$. By abusing notation, we simply refer to it as an atom.

\paragraph{Membership functions.}
An interpretation $\mathcal{I}$ is said to be \emph{consistent} if, 
for every atom 
$p(\mathbf{C};\cdot)$, the truth values over its linguistic labels are compatible according to $\gamma$, that is,
\begin{equation}
\label{eq:consistent}
\left(\ \mathcal{I}\bigl(p(\mathbf{C};\ell)\bigr)\ \right)_{\ell \in \Lambda_p}
\in\gamma_p
\end{equation}

Equation \eqref{eq:consistent} therefore requires that truth degrees of $p(\mathbf{C},\cdot)$ according to $\mathcal{I}$ belong to the admissible region $\gamma_p$ (or they are all zero---recall that $\mathbf{0}^{\Lambda_p}\in \gamma$, if this atom is false at all).
Note that we do not allow arbitrary and independent truth values to individual labels, because membership functions carry out
semantic information. Typically, e.g., it is impossible that an atom corresponding to some physical measure has a high membership value for both the label ``fast'' and the label ``slow''.

In our language, however, the membership functions encoded in $\gamma$ play a much more central role than mere consistency checking: they are directly involved in the inference process, since once a truth value for a predicate $p$ is derived for a label, the profile $\gamma_p$ can  propagate consistent truth values to other labels (line \ref{line:gamma} of Algorithm \ref{algo:leastmodel}).
More precisely, if $\mathcal{I}(p(\mathbf{C};\ell)) = v$, then the profile $\gamma_p$ determines a set of admissible tuples in $[0,1]^{\Lambda_p}$ containing $v$ as the component for $\ell$, and the inference step selects (e.g., minimally) a tuple $\mathbf{t}\in\gamma_p$ such that $\mathbf{t}(\ell)\geq v$, assigning $\mathcal{I}(p(\mathbf{C};\ell')) \geq \mathbf{t}(\ell')$ for every other label $\ell' \in \Lambda_p$.

By this mechanism, partial evidence derived for a single label can be extended to a semantically coherent interpretation over all labels, preserving the intended qualitative relationships encoded in $\gamma_p$ and avoiding arbitrary or inconsistent assignments.

This feature can also be exploited to allow a single predicate $p$ to exhibit both a fuzzy and a crisp behavior. For instance, $p$ can be equipped with two labels, e.g., \textit{fuzzy} and \textit{crisp}, each associated with a different membership function: a continuous increasing function (e.g., linear or even the identity on $[0,1]$) for \textit{fuzzy}, and a $0$-$1$ step function at a fixed cutoff value for \textit{crisp} (see Figure \ref{fig:memb_fuzzycrisp}). In this setting, the label \textit{fuzzy} is used in the head of fuzzy rules to derive graded truth values, while the $\gamma$-propagation mechanism automatically induces a corresponding Boolean value for $p(\cdot;\textit{crisp})$, which will be either $1$ if the threshold degree exceeds the given cutoff, or $0$ otherwise. 

By construction, the label \textit{crisp} thus exhibits a crisp behavior, which can be seamlessly used in subsequent (standard ASP) rule layers requiring Boolean semantics. This pattern is indeed exploited in the program of Figures \ref{fig:p_control}, \ref{fig:p_control2} and \ref{fig:p_control3}, where predicates such as \texttt{discard}, \texttt{discount}, and \texttt{repair} are first derived fuzzily and then used within a purely crisp decision component, by means of their crisp (binary valued) labels.

\paragraph{Models and Answer Sets.}
For a rule $r$, $\mathcal{I}(B(r))$ is obtained by applying the T-norm $\tnorm_r$ to the interpretation $\mathcal{I}(b_i)$ of all literals $b_i\in B(r)$. Recall that $\tnorm_r$ is associative and commutative. Then, $\mathcal{I}(R_h)$ denotes the application of the co-norm associated with $h$ to the interpretation of all body rules defining $h$, that is,
$\mathcal{I}(R_h) = \mathcal{S}_h \{\ \mathcal{I}(B(r)\mid r \in R_h\ \}$.

A \emph{model} of a fuzzy program $\Pi$ is a consistent interpretation $\mathcal{I}$ (in the sense of Equation \eqref{eq:consistent}) such that all rules in $\Pi$ are satisfied.
That is, for each head atom $h$, it holds that
\begin{equation}
\label{eq:model}
\mathcal{I}(h) \geq \mathcal{I}(R_h)
\end{equation}
We write $\mathcal{I}\models\Pi$ to denote that $\mathcal{I}$ is a model of $\Pi$.

\begin{definition}[Minimal model]
An interpretation $\mathcal{I}$ for a program $\Pi$ is said to be \emph{minimal} w.r.t. a given set of interpretations $W$, if there exists no other interpretation $\mathcal{I}'\in W$ ($\mathcal{I}'\neq \mathcal{I}$), such that
 $\mathcal{I}'(\bar p) \leq \mathcal{I}(\bar p)$, for every ground atom $\bar p$.

 An interpretation $\mathcal{I}$ is called a minimal model of $\Pi$, if it is minimal w.r.t. to the set of all models of $\Pi$.
\end{definition}

It is easy to see that a program may have many minimal models, even infinitely many, or no model at all.
However, we are interested in those models that are maximally supported by the program rules.

\begin{definition}[Tight Interpretations]
Let $\mathcal{I}$ and $\mathcal{I'}$ be interpretations of a fuzzy program $\Pi$. 
We say that $\mathcal{I}$ is \emph{tight} w.r.t. $\mathcal{I'}$ if,
for every (unlabeled) atom $p(\mathbf{C};\cdot)$, there is a maximal set of critical labels, whose membership values are the lowest possible to satisfy their defining rules w.r.t. the given $\mathcal{I'}$.
That is, for every such a label $\ell^*$, said $\bar p = p(\mathbf{C};\ell^*)$, we have $\mathcal{I}(\bar p) = \mathcal{I'}(R_{\bar p})$.
If $\mathcal{I}=\mathcal{I'}$, we just say it is a tight interpretation.
\end{definition}

Tight minimal models contain no superfluous truth information: decreasing the truth (membership) value of a label necessarily destroys model-hood (either by violating consistency or by failing to satisfy some rule).
They are maximally informative solutions compatible with both the rules of the program and the semantic constraints imposed by $\gamma$.
If a program has a unique tight minimal model, that model is called the \emph{least model} of the program.

Given an interpretation $\mathcal{I}$ of a fuzzy program $\Pi$, define the \emph{reduct} $\Pi^{\mathcal{I}}$ of $\Pi$ with respect to $\mathcal{I}$ as the set of rules $Pi^{\mathcal{I}}=\{\,r^{\mathcal{I}} \mid r\in\Pi\,\}$,
where each rule $r^{\mathcal{I}}$ is obtained from $r$ by replacing every negated body atom $\neg_{\scriptscriptstyle \mathcal{N}} b$ with 
a (positive) atom $const(\mathcal{N}(\mathcal{I}(b));\ell_c)$.
That is, we assume the truth values of negative literals are correct.
 By construction, $\Pi^{\mathcal{I}}$ is a positive fuzzy program.

\begin{definition}[Fuzzy answer set]
An interpretation $\mathcal{I}$ is an \emph{answer set} of $\Pi$ if $\mathcal{I}$ is a \emph{tight minimal model} of the reduct $\Pi^{\mathcal{I}}$.
\end{definition}

\begin{algorithm}[h]
\KwIn{Positive \Pinc\ fuzzy program}
\KwOut{Least model $\mathcal{M}^*$ (or \texttt{NoModel})}

Construct the Herbrand base $\mathcal{B}_\Pi$;\\
Initialize $\mathcal{I} \leftarrow\bot$;\\
\Repeat{$\mathcal{I}'=\mathcal{I}$}{
  \tcp{Deterministic one-step operator $\widehat{T}$}
  \ForEach{grounded atom $p(\mathbf{C};\cdot)$}{
    \tcp{rule-induced lower bounds}
    \ForEach{$\ell\in\Lambda_p$}{
      $b_\ell \leftarrow \mathcal{I}(R_{p(\mathbf{C};\ell)})$;
    }
    \tcp{unique tight propagation via $\gamma_p$}
    \label{line:gamma}$A \leftarrow \{\,t\in[a_p,b_p]\mid \gamma_p(t)_\ell \ge b_\ell\ \forall \ell\in\Lambda_p\,\}$;\\
    \If{$A=\emptyset$}{\Return{\texttt{NoModel}}}
    $\alpha \leftarrow \min A$;\\
    \ForEach{$\ell\in\Lambda_p$}{
      $\mathcal I'(p(\mathbf C;\ell)) \leftarrow \gamma_p(\alpha)_\ell$;
    }
  }
  $\mathcal I \leftarrow \mathcal I'$;
}
\Return{$\mathcal M^* \leftarrow \mathcal I$;}
\caption{Fixpoint iteration.}
\label{algo:leastmodel}
\end{algorithm}


Let $\mathcal{I}$ be an interpretation for a program $\Pi$. 
Let $W_\mathcal{I}$ be the set of interpretations $\mathcal{I}'$ that satisfy all rules according to $\mathcal{I}$, that is, for every atom $\bar h$,
$\mathcal{I}'(\bar h)\geq \mathcal{I}(R_{\bar h})$.

We define the \emph{immediate consequence operator}  as the mapping 
$\TT: 
[0,1]^{\mathcal{B}_\Pi}\rightarrow\mathcal{P}\bigl([0,1]^{\mathcal{B}_\Pi}\bigr)$
such that $\TT(\mathcal{I})$ is the set of the tight interpretations in $W_\mathcal{I}$ that are minimal w.r.t. $W_\mathcal{I}$.
Moreover, for a set of interpretations 
$\mathcal{\mathbf{I}}$,
define $\TT(\mathcal{\mathbf{I}})= \{ \TT(I) \mid \mathcal{I}\in \mathcal{\mathbf{\mathcal{I}}} \}$.


\begin{definition}[\Pinc\ programs]
A program $\Pi$ is called \emph{\Pinc} if, for each rule $r\in\Pi$ the component function $\gamma_p(t)_\ell: [a_p,b_p]\to[0,1]$ relative to the predicate $p$ and the linguistic label $\ell$ that appears in $H(r)$
is continuous and (strictly) increasing on $[a_p,b_p]$.
\end{definition}


It can be shown that the $\gamma$-monotonicity property of a program $\Pi$ guarantees the existence of a least model, which can be computed via Algorithm \ref{algo:leastmodel}. In contrast, non-monotonic membership profiles may induce nondeterministic behavior, giving rise to multiple tight minimal models.

Furthermore, as observed in other fuzzy logic-based frameworks \cite{Janssen12} even in the monotonic setting, the inference procedure may fail to terminate in a finite number of steps, due to a lack of convergence to the least model. In the framework considered in this paper, however, this is not an issue because we restrict ourselves to a fragment that guarantees polynomial time evaluation (for programs with stratified negation).

\section{Case Study: Industrial Quality Control}
In this section, we present a program $\Pi_{control}$ for a case study arising from a collaboration with an industrial partner, aimed at analyzing images acquired from a screws production line. The goal  is threefold: (i) to detect the presence of damages, (ii) to assess their severity, and (iii) to determine the appropriate action for defective items. As a benchmark, we rely on the well-known MVTec dataset~ \cite{bergmann2021mvtec}, which contains images of objects categorized as normal or defective, with various types of damages.

The implemented system can be conceptually divided into two layers: the first, sub-symbolic, is based on machine learning algorithms that provide the values for the membership functions of the input (EDB) predicates of the symbolic layer. The latter, based on fuzzy reasoning, logically combines this information through the program $\Pi_{control}$, shown in Figures~\ref{fig:p_control}, \ref{fig:p_control2}, and \ref{fig:p_control3}. to obtain the desired results according to the (vague) indications of domain experts.

Note that $\Pi_{control}$ has two modules: the first one, which defines the fuzzy predicates linked to the sub-symbolic methods that process input images, identifies anomalous products and extracts relevant linguistic features describing the detected damages. The second one consists of crisp rules that, based on the inferred qualitative descriptions, support decision-making about the handling of defective items. 
In fact, the proposed language allows us to seamlessly combine rules and predicates with fuzzy values with rules having crisp truth values, which behave exactly as in standard ASP programs.

Each item is defined by a tuple $(X,I)$, where each instance of $X$ is a unique identifier associated with an image tensor $I$. The information is enriched by a neural-generated anomaly heatmap $H$ linked to the image $I$. The anomaly heatmap $H$ highlights pixels corresponding to potential surface defects. This is described by the two facts  $\mathtt{object}(X,I)$ and $\mathtt{anomaly}(I,H)$. Note that these facts are encoded by crisp predicates, indeed we omit linguistic labels in these predicates (formally, we should use a dummy label, say \textit{true}, with a binary membership function).

\begin{figure}
    \centering
    \includegraphics[width=0.32\textwidth]{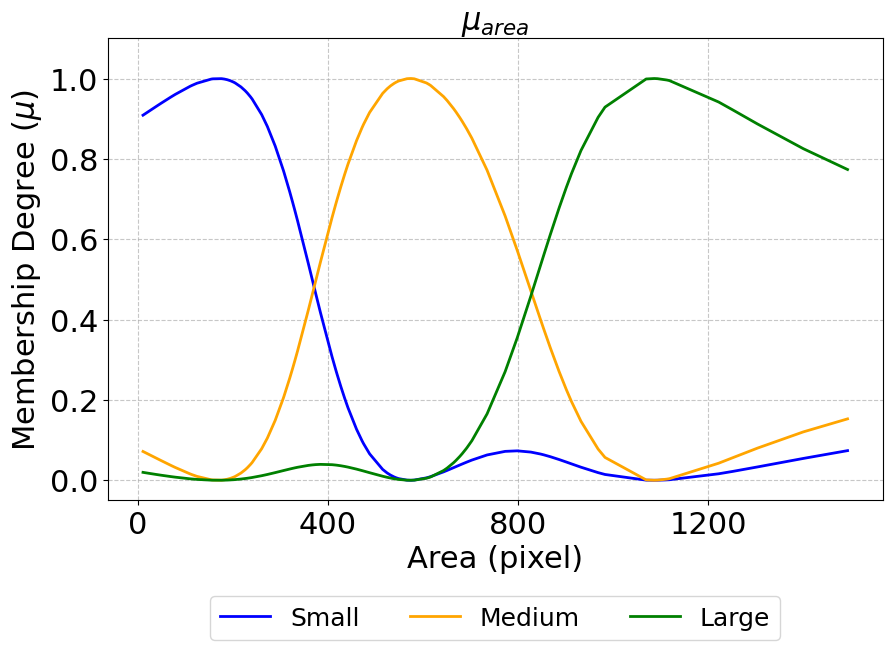}
    \includegraphics[width=0.32\textwidth]{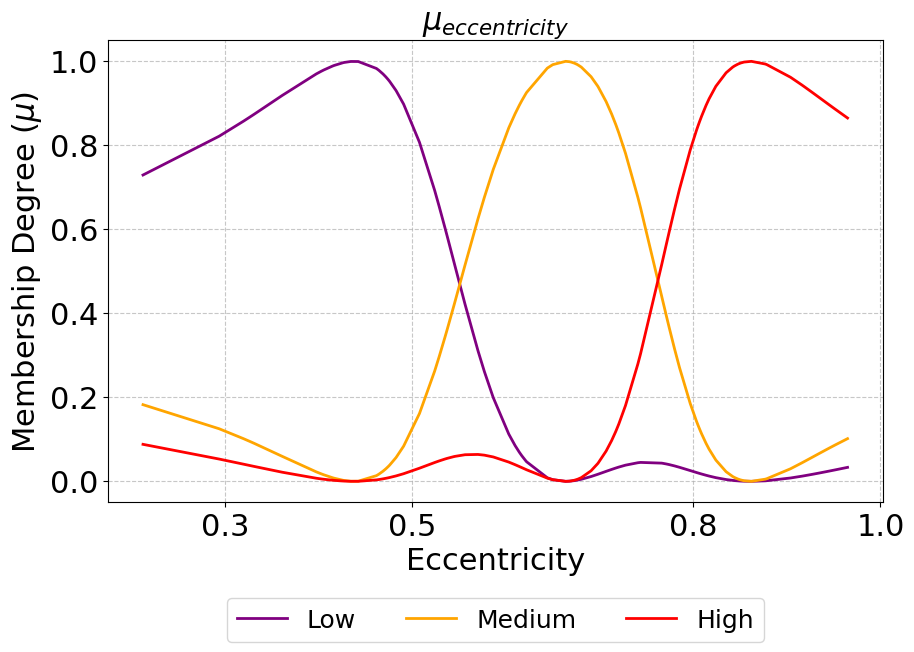}
    \includegraphics[width=0.32\textwidth]{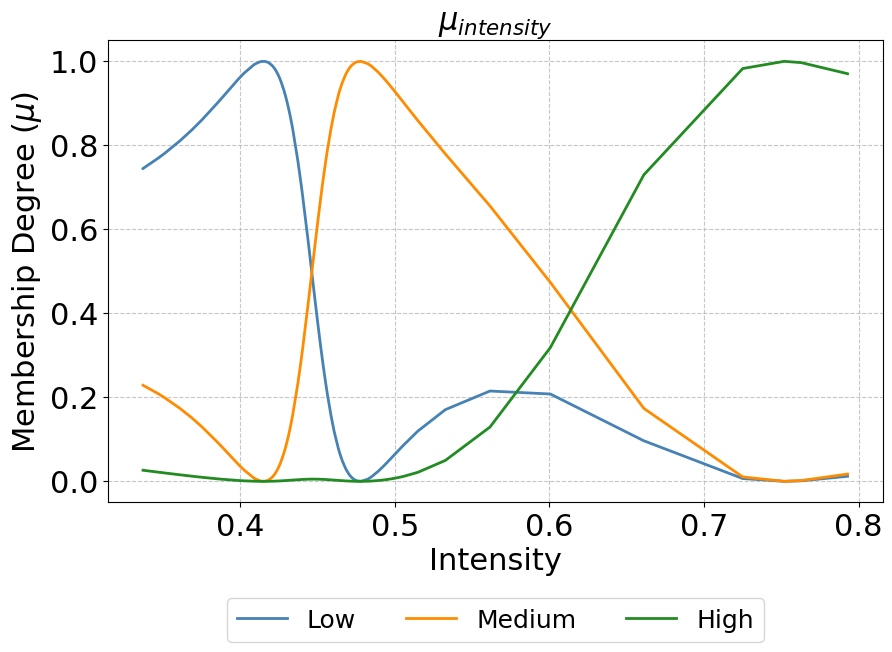}
    \caption{Membership functions of the predicates \texttt{area\_dimension}, \texttt{eccentricity}, and \texttt{color\_intensity}, obtained from sub-simbolic components.}
    \label{fig:mf_kmeans}
\end{figure}

\noindent
The sub-symbolic layer combines: (i) AE-XAD \cite{angiulli2025explaining}, a recent explainable anomaly detection method based on autoencoders, and (ii) three instances of fuzzy $k$-means clustering \cite{nascimento2000fuzzy}. The adoption of AE-XAD is primarily motivated by its ability to produce interpretable anomaly heatmaps that highlight specific pixels corresponding to potential surface defects, providing localized and explainable visual evidence. Furthermore, we rely on fuzzy $k$-means to characterize each detected damage in terms of area dimension, shape eccentricity, color intensity, distance from the center of the image, and solidity. The choice of a fuzzy clustering approach is crucial to manage the ambiguity of irregular shapes and physical defects. Instead of forcing a rigid threshold, it provides graded truth degrees that naturally define the membership functions of the input EDB predicates (\texttt{area\_dimension}, \texttt{eccentricity}, \texttt{color\_intensity}, \texttt{distance}, and \texttt{solidity}), preserving the vagueness of the domain. The linguistic labels of these predicates are reported in the first block of Table \ref{tab:ExampleLabels}, and the corresponding membership functions are displayed in Figure \ref{fig:mf_kmeans}.

\noindent
It is important to note that the raw truth degrees produced by sub-symbolic components (e.g., fuzzy $k$-means) may occasionally exhibit semantically unintuitive behaviors, such as assigning non-zero degrees to both \textit{small} and \textit{large} without a clearly dominant \textit{medium} value, as observed for the predicates \texttt{area\_dimension}, \texttt{eccentricity}, and \texttt{color\_intensity} in the program $\Pi_{control}$ (see Figure \ref{fig:mf_kmeans}).
Such phenomena are intrinsic to membership functions derived from machine learning systems. However, as we show later, the symbolic fuzzy component of our framework is robust to these irregularities: by enforcing semantic constraints and leveraging well-designed rules, it mitigates the imprecision of sub-symbolic outputs, yielding more coherent, precise, and interpretable results.

\begin{figure}[h]
    \centering
    \begin{subfigure}[t]{0.32\linewidth}
        \centering
        \includegraphics[width=\linewidth]{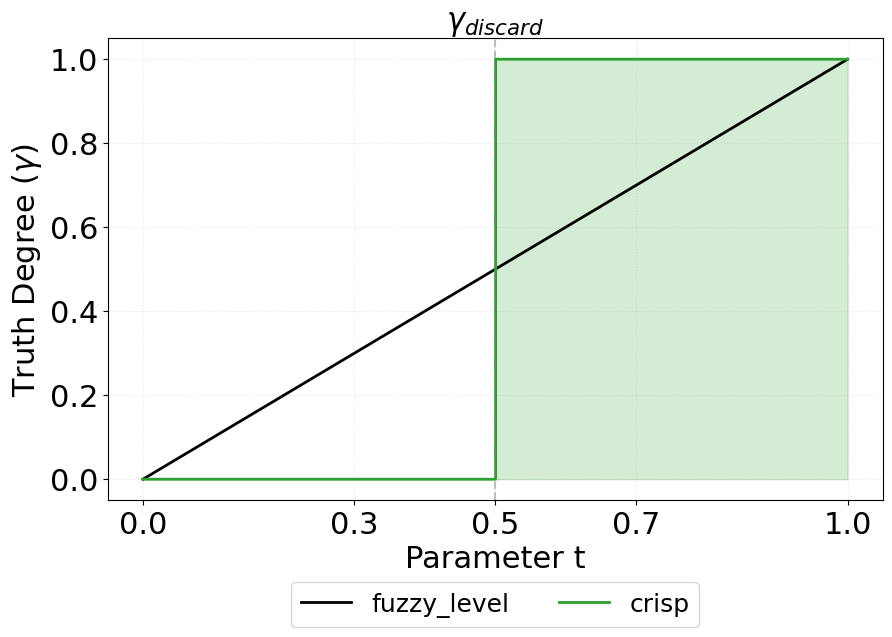}
        \caption{Predicates \texttt{discard} and \texttt{repair}.}
        \label{fig:mf_discard}
    \end{subfigure}
    \begin{subfigure}[t]{0.32\linewidth}
        \centering
        \includegraphics[width=\linewidth]{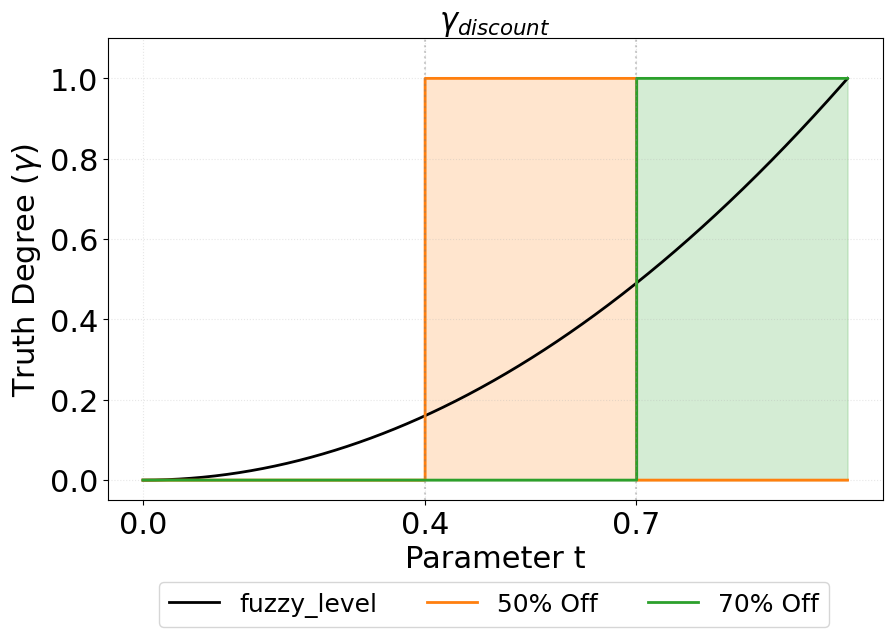}
        \caption{Predicate \texttt{discount}.}
        \label{fig:mf_discount}
    \end{subfigure}
    \begin{subfigure}[t]{0.32\linewidth}
        \centering
        \includegraphics[width=\linewidth]{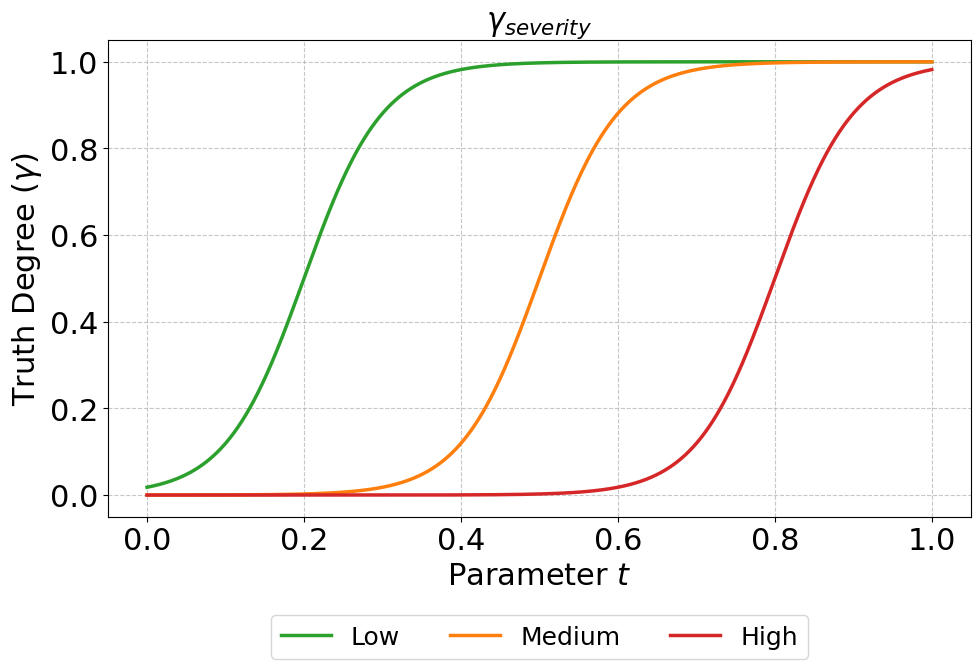}
        \caption{Predicates \texttt{severity} and \texttt{defect}.}
        \label{fig:mf_severity}
    \end{subfigure}
    \caption{Membership functions for the predicates \texttt{discard}, \texttt{discount}, \texttt{repair}, and \texttt{severity}. The predicates \texttt{discard} and \texttt{repair} are equipped with a simple fuzzy-to-crisp pattern, with a continuous label (\textit{fuzzy}) and a corresponding Boolean label (\textit{crisp}) obtained via a single cutoff. The predicate \texttt{discount} instead adopts a multi-threshold scheme, where three distinct step functions map the fuzzy degree to crisp decisions corresponding to different discount levels (50\% and 70\%). The predicates \texttt{severity} and \texttt{defect} are entirely fuzzy (no crisp labels) with $\gamma$-monotonic truth profiles.}
    \label{fig:memb_fuzzycrisp}
\end{figure}

\paragraph{$\Pi_{control}$ Rules.}
The symbolic layer of $\Pi_{control}$ employs a set of rules to evaluate anomaly severity, classify defects, and determine the optimal handling strategy. Initially, the framework assesses the overall severity of detected anomalies by evaluating their geometric and visual features, assigning, by means of the rules \eqref{rule:sev1} and \eqref{rule:sev2}, high severity to anomalies with significant area, high eccentricity, or elevated color intensity, while classifying small, faint anomalies as low severity, through rule \eqref{rule:sev3}. At the same level, the rules \eqref{rule:funct1}, \eqref{rule:funct2}, and \eqref{rule:aest} semantically distinguish between functional and aesthetic defects based on the physical traits of the screws. Functional defects are typically characterized by low solidity or specific combinations of small size, low color intensity, and high distance from the center, whereas aesthetic damages are identified through high solidity, low distance, and medium color intensity. 
In particular, note that the parts farthest from the center are precisely the head and the tip of the screw, whose possible absence in the image is clearly associated with defects that compromise the functionality of the screw under examination.

Based on these inferred classifications and their corresponding truth degrees, the subsequent block of rules infers the appropriate handling actions. Objects are flagged for discarding if they exhibit severe functional defects (rules \eqref{rule:discard1} and \eqref{rule:discard2}), or multiple critical aesthetic anomalies (rule \eqref{rule:discard3}). On the other hand,  rule \eqref{rule:discount} provides a recommendation for discounting of objects with high-severity aesthetic damages. Repair strategies, encoded in  rules \eqref{rule:repair_self} and \eqref{rule:repair_ext}, are assigned based on the defect profile: internal repairs target minor aesthetic flaws lacking severe functional implications, while external repairs are reserved for minor functional defects requiring specialized processing.
That is, minor damages can be handled internally by the company, whereas more severe but still repairable damages must be delegated to specialized external companies. Finally, items designated for external repair that do not meet the crisp discard criteria are automatically routed to an available repair company (rule \eqref{rule:send}) and the price of the items designated for discount is updated (rules \eqref{rule:sell1}, \eqref{rule:sell2}, \eqref{rule:sell3}).

We point out that these final rules behave precisely as in standard ASP, because all their body predicates use only binary (crisp) step membership functions, and thus will be either (entirely) true or (entirely) false in any interpretation of the program.
To see how this works, consider for instance the predicate \texttt{discard}: the compatibility $\gamma$ of membership functions (see Figure~\ref{fig:memb_fuzzycrisp}) implies that any \textit{fuzzy\_level} value exceeding a certain threshold automatically sets the value $1$ (entirely true) for the step membership function we have called \textit{crisp}, which is then used in the subsequent rules (e.g., rule \eqref{rule:send}).
The same happens for predicates \texttt{repair} and \texttt{discount}.

Moreover, it is important to note that some rules of $\Pi_{control}$ exhibit peculiar labels containing the symbols $\geq$ and $\leq$. For example, in the rule \eqref{rule:funct2}, the predicate \texttt{color\_intensity} is used with the label $\leq medium$.
Intuitively, this expression extends the linguistic notion of \textit{at least medium}.
Formally, these labels are in fact additional membership functions
that are usually obtained by combining other membership functions of the same predicate.
In the considered program, this combination is performed by means of the probabilistic sum operator, e.g., $\mathit{p}(;\ge \mathit{medium}) = \mathit{p}(;\mathit{medium}) + \mathit{p}(;\mathit{high}) - \mathit{p}(;\mathit{medium}) \cdot \mathit{p}(;\mathit{high})$.

\begin{table}[h]
    \centering
    \begin{tabular}{|c|c|}
        \hline 
        \textbf{Predicate} & \textbf{Linguistic Labels}\\ 
        \hline
        \texttt{area\_dimension} & $\{small, medium, large\}$ \\
         \texttt{eccentricity} & $\{low, medium, high\}$ \\
         \texttt{color\_intensity} & $\{low, medium, high\}$\\
         \texttt{solidity}  & $\{low, medium, high\}$\\
         \texttt{distance} &  $\{low, medium, high\}$ \\
        \hline
         \texttt{severity} & $\{low, medium, high\}$\\
         \texttt{defect} & $\{low, medium, high\}$\\
         \texttt{discard} & $\{fuzzy\_level, crisp\}$\\
         \texttt{discount} & $\{fuzzy\_level, disc50, disc70\}$\\
         \texttt{repair} & $\{fuzzy\_level, crisp\}$\\
        \hline
    \end{tabular}
    \caption{Corresponding set of labels for each predicate.}
    \label{tab:ExampleLabels}
\end{table}

\begin{figure}[h!]
\centering
{\footnotesize
\hrule
\textbf{$\Pi_{control}$: Industrial Quality Control Program}
\hrule
\begin{align}
\refstepcounter{proprule}\label{rule:sev1}
\tag{\theproprule}
    \begin{split}
        \mathtt{severity}(X, H; high) \leftarrow & \: \mathtt{object}(X, I) \odot_{\min} \mathtt{anomaly}(I, H)  \odot_{\min} \\ & \mathtt{area\_dimension}(I, H; large) \odot_{\min} \\ & \mathtt{eccentricity}(I, H; high)
    \end{split} \\
\refstepcounter{proprule}\label{rule:sev2}
\tag{\theproprule}
    \begin{split}
        \mathtt{severity}(X, H; high) \leftarrow & \: \mathtt{object}(X, I) \odot_{\min} \mathtt{anomaly}(I, H) \ \odot_{\min} \\ & \mathtt{color\_intensity}(I, H;\geq medium ) \ \odot_{\min} \\ & \mathtt{eccentricity}(I, H; high)
    \end{split}\\
\refstepcounter{proprule}\label{rule:sev3}
\tag{\theproprule}
    \begin{split}
        \mathtt{severity}(X, H; low )\leftarrow & \:\mathtt{ object}(X, I) \ \odot_{\min} \mathtt{anomaly}(I, H) \ \odot_{\min} \\ & \mathtt{color\_intensity}(I, H; \leq medium) \ \odot_{\min} \\ & \mathtt{area\_dimension}(I, H; low)
    \end{split}\\
\refstepcounter{proprule}\label{rule:funct1}
    \tag{\theproprule}
    \begin{split}
        \mathtt{defect}(I, H, functional; high) \leftarrow & \: \mathtt{anomaly}(I, H)  \odot_{\min} \\ & \mathtt{area\_dimension}(I, H; small) \ \odot_{\min} \\ & \mathtt{distance}(I, H; high) \odot_{\min} \\ & \mathtt{color\_intensity}(I, H; low)
    \end{split} \\
\refstepcounter{proprule}\label{rule:funct2}
\tag{\theproprule}
    \begin{split}
    \mathtt{defect}(I, H, functional; high) \leftarrow & \: \mathtt{anomaly}(I, H) \odot_{\min} \\ & \mathtt{color\_intensity}(I, H; \leq medium) \odot_{\min} \\ & \mathtt{solidity}(I, H; low)
        \odot_{\min} \\ & \mathtt{distance}(I, H; medium) 
        \end{split} 
        \\
\refstepcounter{proprule}\label{rule:aest}
\tag{\theproprule} \begin{split} 
     \mathtt{defect}(I, H, aesthetic; high)\leftarrow & \: \mathtt{anomaly}(I, H) \odot_{\min} \\ & \mathtt{color\_intensity}(I, H; medium) \ \odot_{\min} \notag \\ & \mathtt{solidity}(I, H; high)
        \odot_{\min} \\ & \mathtt{distance}(I, H; low) 
        \end{split}
\end{align}
\hrule}
\caption{The Fuzzy Datalog program $\Pi_{control}$. Defect identification.}

\label{fig:p_control}
\end{figure}

\begin{figure}[h!]
\centering
{\footnotesize
\hrule
\textbf{$\Pi_{control}$: Industrial Quality Control Program}
\hrule
\begin{align}
    \refstepcounter{proprule}\label{rule:discard1}
\tag{\theproprule} \begin{split} 
    \mathtt{discard}(X; fuzzy\_level) \leftarrow & \: \mathtt{object}(X, I) \odot_{\min} \mathtt{anomaly}(I, H) \ \odot_{\min} \\ &
    \mathtt{defect}(I, H, functional; high) \odot_{\min} \\ & \mathtt{severity}(X, H; medium) \end{split} \\
    \refstepcounter{proprule}\label{rule:discard2}
\tag{\theproprule} \begin{split}
    \mathtt{discard}(X; fuzzy\_level) \leftarrow & \: \mathtt{object}(X, I) \odot_{\min} \mathtt{anomaly}(I, H) \ \odot_{\min} \\ &
    \mathtt{defect}(I, H, functional; medium) \odot_{\min} \\ & \mathtt{severity}(X, H; high) \end{split} \\
    \refstepcounter{proprule}\label{rule:discard3}
\tag{\theproprule} \begin{split} 
    \mathtt{discard}(X; fuzzy\_level) \leftarrow & \: \mathtt{object}(X, I) \odot_{\min} \mathtt{anomaly}(I, H_1) \odot_{\min} \\ &  \mathtt{anomaly}(I, H_2) \odot_{\min} \\ &
    \mathtt{defect}(I, H_1, aesthetic; high) \ \odot_{\min} \\ & \mathtt{defect}(I, H_2, aesthetic; high) \odot_{\min} \\ & H_1 \neq H_2 \end{split}\\
    \refstepcounter{proprule}\label{rule:discount}
\tag{\theproprule} \begin{split} 
    \mathtt{discount}(X; fuzzy\_level) \leftarrow & \: \mathtt{object}(X, I) \odot_{\min} \mathtt{anomaly}(I, H) \ \odot_{\min} \\ &
    \mathtt{defect}(I, H, aesthetic; high) \odot_{\min} \\ & \mathtt{severity}(X, H; high) \end{split}\\
    \refstepcounter{proprule}\label{rule:repair_self}
\tag{\theproprule} \begin{split}
    \mathtt{repair}(X, self; fuzzy\_level) \leftarrow & \: \mathtt{object}(X, I) \odot_{\min} \mathtt{anomaly}(I, H) \ \odot_{\min} \\ &
    \mathtt{defect}(I, H, aesthetic; \leq medium) \ \odot_{\min} \\ & \neg_{\scriptscriptstyle\textit{\L{}}}
    \mathtt{defect}(I, H, functional; high) \end{split}\\
    \refstepcounter{proprule}\label{rule:repair_ext}
\tag{\theproprule} \begin{split}
    \mathtt{repair}(X, external; fuzzy\_level) \leftarrow & \: \mathtt{object}(X, I) \odot_{\min} \mathtt{anomaly}(I, H) \ \odot_{\min} \\ &
    \mathtt{defect}(I, H, functional; low) \ \odot_{\min} \\ & \neg_{\scriptscriptstyle\textit{\L{}}}
    \mathtt{defect}(I, H, aesthetic; low)\end{split} 
\end{align}
\hrule}
\caption{The Fuzzy Datalog program $\Pi_{control}$. Fuzzy possible evaluation.}

\label{fig:p_control2}
\end{figure}

\begin{figure}[h!]
\centering
{\footnotesize
\hrule
\textbf{$\Pi_{control}$: Industrial Quality Control Program}
\hrule
\begin{align}
    \refstepcounter{proprule}\label{rule:send}
\tag{\theproprule} \begin{split}
    \mathtt{send}(X, A) \leftarrow & \: \mathtt{object}(X, I) \odot_{\min}
    \mathtt{repair}(X, external; crisp) \ \odot_{\min} \\ &   \neg_{\scriptscriptstyle\textit{\L{}}}
    \mathtt{discard}(X; crisp) \odot_{\min} \mathtt{available\_company}(A) \end{split}  \\
\refstepcounter{proprule}\label{rule:sell1}
\tag{\theproprule} \begin{split}
    \mathtt{sell}(X, P) \leftarrow & \: \mathtt{object}(X, I) \odot_{\min} \mathtt{original\_price}(X, P_0) \ \odot_{\min} \\ &  \mathtt{discount}(X; disc50) \ \odot_{\min} P=P_0*0.5 \ \odot_{\min} \\ & \neg_{\scriptscriptstyle\textit{\L{}}} \mathtt{discard}(X; crisp)\end{split}\\
\refstepcounter{proprule}\label{rule:sell2}
\tag{\theproprule} \begin{split}
    \mathtt{sell}(X, P) \leftarrow & \: \mathtt{object}(X, I) \odot_{\min} \mathtt{original\_price}(X, P_0) \ \odot_{\min} \\ &  \mathtt{discount}(X; disc70) \ \odot_{\min} P=P_0*0.7   \ \odot_{\min} \\ & \neg_{\scriptscriptstyle\textit{\L{}}} \mathtt{discard}(X; crisp) \end{split}\\
\refstepcounter{proprule}\label{rule:sell3}
\tag{\theproprule} \begin{split}
    \mathtt{sell}(X, P) \leftarrow & \: \mathtt{object}(X, I) \odot_{\min} \mathtt{original\_price}(X, P) \odot_{\min} \\ & \neg_{\scriptscriptstyle\textit{\L{}}} \mathtt{discard}(X; crisp) \odot_{\min} \neg_{\scriptscriptstyle\textit{\L{}}} \mathtt{discount}(X; disc50) \odot_{\min} \\ & \neg_{\scriptscriptstyle\textit{\L{}}} \mathtt{discount}(X; disc70)  \end{split}
\end{align}
\hrule}
\caption{The Fuzzy Datalog program $\Pi_{control}$. Crisp decision.}

\label{fig:p_control3}
\end{figure}

\subsection{Examples of Evaluation}

\begin{figure}[h]
    \centering
    \begin{subfigure}{0.48\linewidth}
        \centering
        \includegraphics[width=\linewidth]{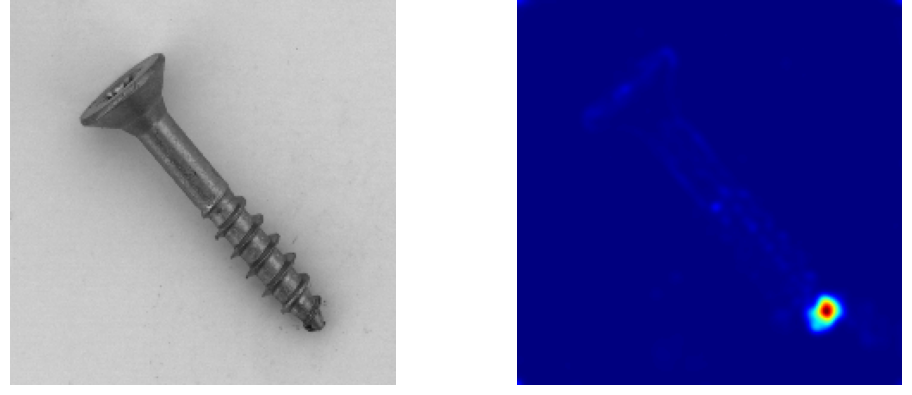}
        \caption{Manipulated Front Anomaly.}
        \label{fig:mf}
    \end{subfigure}
    \hfill
    \begin{subfigure}{0.48\linewidth}
        \centering
        \includegraphics[width=\linewidth]{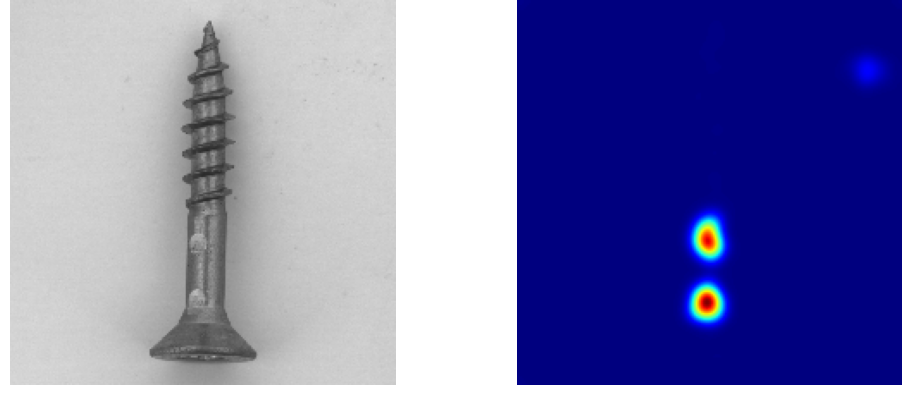}
        \caption{Two Scratch Anomalies.}
        \label{fig:scr}
    \end{subfigure}
    \caption{Two input examples with corresponding AE-XAD anomaly explanations.}
    \label{fig:anomalies}
\end{figure}

In this section, we present the derivation for two specific instances (displayed in Figure \ref{fig:anomalies}). In this application we explicitly evaluate the t-norm and s-norm using the $min$ and $max$ operators respectively, for every clause in the extended $\Pi_{control}$ program.
The $min$ operator ensures that when aggregating multiple conditions in a rule body (e.g., determining if a defect is highly severe), the resulting truth degree is strictly bounded by the weakest piece of evidence. Conversely, the $max$ operator (s-norm) guarantees that when multiple, independent rules derive the same concept for an item, the strongest available evidence dictates the final classification. This prevents severe defects from being smoothed out or averaged down by less critical features.

In real-world scenarios, the transition between qualitative concepts—such as moving from a \textit{(at least) medium} to a \textit{(at least) high} severity—is rarely abrupt. We adopt a \emph{smooth approach}, as shown in Figure \ref{fig:mf_severity}, which provides continuous transitions for such notions. 

We define our admissible monotonic truth profiles $\gamma$ by mapping each ordered label $\ell_i$ to a shifted logistic sigmoid \citep{huybrechs2024sigmoid}:
$$
f(t; k, c_i) = \sigma\big(k(t - c_i)\big) = \frac{1}{1 + e^{-k(t - c_i)}}
$$ 
where $k>0$ governs the steepness and $c_i \in (0,1)$ determines the threshold center. We fix $k=20$, allowing the closed-form inverse derivation of the parameter value $t = c_i + \frac{1}{20} \ln\left(\frac{y}{1-y}\right)$.

For a predicate like \texttt{severity} with three ordinal labels ($\mathit{low}$, $\mathit{medium}$, $\mathit{high}$), the profile maps a curve to each threshold using monotonically increasing centers $c_1=0.20$, $c_2=0.50$, and $c_3=0.80$:
$$
\gamma_{\mathit{severity}}(t)= \big( f(t; 20, 0.20),\; f(t; 20, 0.50),\; f(t; 20, 0.80) \big)
$$

The final decision stage of the pipeline is modeled through the predicates \texttt{discard}, \texttt{discount}, and \texttt{repair}.

As explained above, these predicates are equipped with linguistic labels that enable a transition from fuzzy evaluations to crisp behaviors by means of their associated membership functions depicted in Figure \ref{fig:memb_fuzzycrisp}.
In more detail, the predicates \texttt{discard} and \texttt{repair} are defined with two labels, \textit{fuzzy\_level} and \textit{crisp}: the former is associated with a continuous increasing membership function (in our case, linear in $[0,1]$), while the latter is defined via a step function with a cutoff, fixed in this example at $0.5$. This means that a fuzzy degree assigned to \texttt{discard}$(X;\textit{fuzzy\_level})$ (respectively \texttt{repair}$(X;\textit{fuzzy\_level})$) is turned into a crisp value for \texttt{discard}$(X;\textit{crisp})$ (respectively \texttt{repair}$(X;\textit{crisp})$), which is $1$ if the degree is at least $0.5$, and $0$ otherwise.

The predicate \texttt{discount} follows a similar approach, but it is associated with two crisp labels corresponding to different discount levels, namely \textit{disc50} and \textit{disc70}. Each of these labels is defined by a step membership function with thresholds set to $0.4$ and $0.7$, respectively. As a result, the fuzzy degree computed for \texttt{discount}$(X;\textit{fuzzy\_level})$ is mapped, via the $\gamma$-propagation mechanism, to either crisp discount decision depending on which thresholds are exceeded.

Note that the threshold values associated with the fuzzy-to-crisp predicates are meant to capture a specific decision policy and are application-dependent design parameters. Future work will investigate data-driven approaches for automatically learning these thresholds, in addition to the shape and parameters of the underlying membership functions.

\paragraph{Instance 1: Manipulated Front Anomaly.}

Let $X_{1}$ be the item featuring a single manipulated front anomaly $H$ (as seen in Figure \ref{fig:mf}). In this scenario, the geometric evaluation maps the defect to a functional anomaly. 

We focus the explicit derivation on the \texttt{severity} predicate to showcase the continuous latent inference. First, we compute the required compound labels via probabilistic sum ($a \oplus b = a + b - a \cdot b$):
\begin{align*}
    \mathtt{color\_intensity}(X_1, H;\le medium) &= 0.998 \oplus 0.002 = 0.998 \\
    \mathtt{color\_intensity}(X_1, H;\ge medium) &= 0.002 \oplus 0.000 = 0.002
\end{align*}
Next, we evaluate the lower bounds dictated by rules (\ref{rule:sev1}--\ref{rule:sev2}) using the $\min$ t-norm over the geometric memberships:
{\footnotesize\begin{align*}
    \mathtt{severity}(H; high) &\leftarrow \max \big( \min(0.020, 0.074), \min(0.002, 0.074) \big) = 0.020 \quad \text{(\ref{rule:sev1}, \ref{rule:sev2})}\\
    \mathtt{severity}(H; low) &\leftarrow \min(0.998, 0.909) = 0.909 \quad \text{(\ref{rule:sev3})}
\end{align*}}
With the constraints $\mathit{high} \ge 0.020$ and $\mathit{low} \ge 0.909$, we map the evaluations to the latent state $t$ using the inverse sigmoid. The constraint on $\mathit{high}$ ($c_3=0.80$) imposes $t \ge 0.80 + 0.05 \ln(0.020/0.980) = 0.605$. This is strictly greater than the requirement for $\mathit{low}$ ($t \ge 0.315$). Thus, substituting the strictest bound $t = 0.605$ back into the $\gamma_{\mathit{severity}}$ profile provides the fully coherent semantic state

 $$\mathtt{severity}(H; \cdot) \mbox{ with labels }\{low=0.999, \;medium = 0.892, \; high = 0.020\}. $$

The framework successfully infers a strong $\mathit{medium}$ severity (interpreted as ``at least medium'' because these monotonic function encode a cumulative behavior), despite the absence of an explicit rule, naturally bridging the gap between the $\mathit{low}$ and $\mathit{high}$ evaluations.

Processing the morphological variables through rules (\ref{rule:funct1}--\ref{rule:funct2}) yields a confident classification for a functional defect ($\mathtt{defect}(I, H, functional; high) = 0.909$), primarily driven by the anomaly's localized area and its position at the object's boundaries.

$$ \mathtt{discard}(X_1; fuzzy\_level)) \leftarrow \min(0.909, 0.892) = 0.892 $$
Because of the threshold profile $\gamma_{\mathit{discard}}$ (with cutoff $0.5$), we get a definitive crisp action: $\mathrm{discard}(X_1; \mathit{crisp}) = 1$.

\paragraph{Instance 2: Two Scratch Anomalies (Image \ref{fig:scr}).}

Let $X_{2}$ be an item presenting two candidate anomaly regions identified by the sub-symbolic layer, denoted as $H_A$ (Component 1) and $H_B$ (Component 2) (Figure \ref{fig:scr}).
We evaluate the severity profiles independently for both components. For $H_A$, we compute the aggregated bounds via probabilistic sum ($\mathtt{color\_intensity}(X_2, H_A;\le medium) = 0.043 \oplus 0.953 = 0.955$) and apply rules (\ref{rule:sev1}--\ref{rule:sev3}):
\begin{align*}
    \mathtt{severity}(H_A; high) &\leftarrow\max \big(\min(0.037, 0), \min(0.953, 0) \big) = 0.000 \\
    \mathtt{severity}(H_A; low) &\leftarrow \min(0.955, 0.264) = 0.264
\end{align*}
With bounds $high \ge 0$ and $low \ge 0.264$, the strictest latent state is $t = 0.148$. Substituting this into $\gamma_{\mathit{severity}}$ yields the following profile for the first component $\mathtt{severity}(H_A,\cdot)$: $\{low=0.261, \;medium=0.001, \;high=0.000\}$.
Conversely, for the region $H_B$, the probabilistic sum yields $\mathtt{color\_intensity}(X_2, H_B;\le medium) = 0.056 \oplus 0.938 = 0.941$ and $\mathtt{color\_intensity}(X_2, H_B;\ge medium) = 0.938$. Then,
\begin{align*}
    \mathtt{severity}(H_B; high) &\leftarrow \max \big( \min(0.038, 0.026), \min(0.938, 0.026) \big) = 0.026 \\
    \mathtt{severity}(H_B; low) &\leftarrow \min(0.941, 0.272) = 0.272
\end{align*}
Here, the constraint on $high$ governs the latent inference, forcing $t \ge 0.80 + 0.05 \ln(0.026/0.974) = 0.619$. The semantic state naturally bridges the evaluations:
$$ \mathtt{severity}(H_B,\cdot) \mbox{ with labels } \{low=0.999, \;medium = 0.915, \; high = 0.026\}. $$

Aggregating the truth degrees across the previous rule set, and applying step-function profile $\gamma_{\mathit{discard}}$ yields $\mathtt{discard}(X_2; crisp) = 0$. The framework safely avoids a false-positive discard.

Instead, the presence of a verified single aesthetic anomaly ($H_B$) with a safely bounded severity enables the self-repair policy. Given the latent state of $H_B$, the cumulative label $\mathtt{defect}(H_B, aesthetical; \le medium)$ evaluates to $1.000$. Coupled with the \L{}ukasiewicz negation of functional damages ($1 - 0.023 = 0.977$), we obtain:
$$ \mathtt{repair}(X_2, self; fuzzy\_level) \leftarrow \min(1.000, 0.977) = 0.977$$
Through $\gamma_{\mathit{repair}}$, this yields  $\mathtt{repair}(X_2, self; crisp) = 1$.

The neuro-symbolic pipeline leverages declarative geometric semantics to route each item to the most appropriate and cost-effective recovery procedure. The examples are handled correctly: the first is discarded by the system due to a recognized functional defect, while the second is self-repaired, as expected for scratch anomalies.

The choice of t-norm and s-norm, however, strongly affects the final outcome. As an example, in Instance 1, the product or the \L{}ukasiewicz t-norm would prevent the expected discard decision despite a clearly defective screw, as both degrade the truth of a conjunction as its conditions grow, unlike \textit{min}. Analogously, \textit{max} retains only the strongest supporting rule, whereas additive s-norms may turn several weak signals into a strong conclusion, which is undesirable in this quality control scenario.

The propagation by means of the $\gamma$ function highlights a fundamental theoretical advantage of our framework: the shift from continuous fuzzy reasoning to crisp operational decisions comes easily. Instead of relying on external procedural scripts, ad-hoc defuzzification algorithms, or hardcoded post-processing rules, the binarization may be handled entirely within the logical semantics. 
Ambiguous neural outputs are gracefully absorbed by the continuous logic, naturally ending into deterministic boolean actions without ever breaking the end-to-end logical pipeline.

\section{Conclusion}

In this paper, we presented a case study of recently introduced qualitative fuzzy extension of Answer Set Programming, where predicates are equipped with finite linguistic labels and membership functions.
The prototype leverages the structure of this language to combine a machine learning-based component with a set of symbolic rules to detect and localize damages in images obtained from a production line of screws, and to provide decision-making pattern for damaged items. 
The illustrated examples reveal the capacity of our language to incorporate the extremely effective, although potentially noisy, predictions of machine learning models with the rigor of symbolic reasoning into a formal, declarative pipeline. The association of data-driven membership functions with linguistic labels allows domain experts to encode their knowledge naturally, capturing the inherent vagueness of qualitative concepts without sacrificing logical power.

\bibliographystyle{plain}
\bibliography{biblio}

\end{document}